\documentclass[10pt,twocolumn,letterpaper]{article}

\usepackage{iccv}
\usepackage{times}
\usepackage{epsfig}
\usepackage{graphicx}
\usepackage{amsmath}
\usepackage{amssymb}
\usepackage{booktabs}
\usepackage{multirow}
\usepackage{arydshln}
\usepackage{mathrsfs}
\usepackage{pifont}
\usepackage{xcolor}
\usepackage{authblk}
\usepackage[accsupp]{axessibility}

\usepackage[ruled,vlined]{algorithm2e}

\definecolor{commentcolor}{RGB}{110,154,155}   
\newcommand{\PyComment}[1]{\ttfamily\textcolor{commentcolor}{\# #1}}  
\newcommand{\PyCode}[1]{\ttfamily\textcolor{black}{#1}} 

\newcommand{\cmark}{\ding{51}}
\newcommand{\xmark}{\ding{55}}
\newcommand{\dmark}{\cmark\kern-1.2em\cmark}

\usepackage[pagebackref=true,breaklinks=true,letterpaper=true,colorlinks,bookmarks=false]{hyperref}

\iccvfinalcopy 

\def\iccvPaperID{7277} 

\def\authorBlock{
    $\text{Yeonghwan Song}^{1}$ \qquad
    $\text{Seokwoo Jang}^{1}$ \qquad
    $\text{Dina Katabi}^{2}$ \qquad
    $\text{Jeany Son}^{1}$ \\\vspace{-0.3cm}
    $^{1}\text{AI Graduate School, GIST}$ \qquad
    $^{2}\text{MIT CSAIL}$ \\
    {\tt\small \{yeonghwan.song, jangseokwoo\}@gm.gist.ac.kr} \qquad
    {\tt\small dina@csail.mit.edu} \qquad
    {\tt\small jeany@gist.ac.kr}
}

\begin{document}
\title{Unsupervised Object Localization with Representer Point Selection}
\author{\authorBlock}


\maketitle
\ificcvfinal\thispagestyle{empty}\fi

\begin{abstract}

We propose a novel unsupervised object localization method that allows us to explain the predictions of the model by utilizing self-supervised pre-trained models without additional finetuning.
Existing unsupervised and self-supervised object localization methods often utilize class-agnostic activation maps or self-similarity maps of a pre-trained model.
Although these maps can offer valuable information for localization, their limited ability to explain how the model makes predictions remains challenging.
In this paper, we propose a simple yet effective unsupervised object localization method based on representer point selection, where the predictions of the model can be represented as a linear combination of representer values of training points.
By selecting representer points, which are the most important examples for the model predictions, our model can provide insights into how the model predicts the foreground object by providing relevant examples as well as their importance. 
Our method outperforms the state-of-the-art unsupervised and self-supervised object localization methods on various datasets with significant margins and even outperforms recent weakly supervised and few-shot methods.
Our code is available at: \small{\url{https://github.com/yeonghwansong/UOLwRPS}}

\end{abstract}

\section{Introduction}
\label{sec:intro}

Object localization is one of the most fundamental problems in computer vision which aims to find {a bounding box of a particular object category in a given image.}
Recent object localization methods achieve impressive performances thanks to advances in deep neural networks (DNNs)~\cite{resnet, VIT} and large-scale datasets~\cite{imagenet, CUB}.
Despite their successes, the biggest concern in this task is that collecting datasets with precise box-level annotations is labor-intensive and time-consuming.
Recently, object localization methods with less supervision,~\ie weakly-supervised methods~\cite{HaS, ACoL, cutmix, ADL, SPG, cream, FAM, BAS, fcam}, have been well-studied to mitigate those problems, nevertheless, they require image-level class labels.

\begin{figure}
    \centering
    \includegraphics[width=1.05\linewidth]{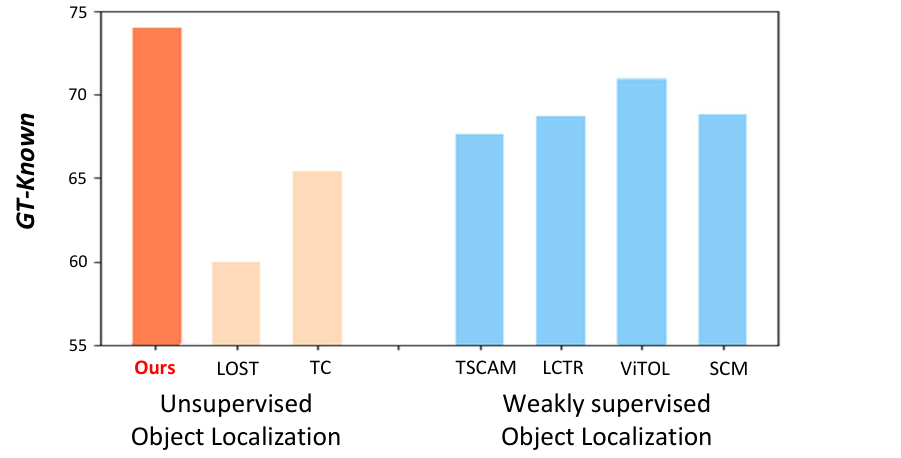}
    \label{fig:graph}
    \vspace{-0.4cm}
    \caption{\textit{GT-Known Loc} scores of various methods with respect to different levels of supervision on ImageNet-1K dataset. All methods use ViT-S~\cite{VIT} as a backbone and TC denotes TokenCut~\cite{tokencut}.}
    \vspace{-0.3cm}
\end{figure}

Beyond learning with less supervision, most recent works~\cite{DDT, SCDA, OLM, Psy, Cocon, Enhance, JointGraph, CCAM, LOST, tokencut} focus on the object localization task using self-supervised or unsupervised learning that does not require any human annotated labels.
These works address the problem of identifying which regions are more likely to contain \textit{the foreground object}, which is a salient object in an image.
In order to discover foreground object regions, several methods~\cite{Psy, Cocon, Enhance, JointGraph} attempt to use the magnitude of feature vectors as a clue for class-agnostic activation maps (CAAM)~\cite{Psy}.
Most of these methods rely heavily on pre-trained models designed for the image classification task.
However transferring these models to the object localization task is challenging, since foreground features are not easily distinguished from the background.
To address this problem, these works have been focusing on learning to discriminate foreground and background representations from the pre-trained model.

Recent approaches for unsupervised object localization face a limitation in terms of the explainability of the predictions. 
{DNNs} are powerful predictors across various domains, but their complicated structure often makes them black-box functions. 
Class activation maps (CAM){~\cite{CAM}} are frequently used to provide visually interpretable information about a specific class the model has learned, but a detailed explanation of how the model makes its predictions remains unclear.
This limitation becomes even more severe for CAAM, which is commonly used in self-supervised object localization methods~\cite{Psy, Cocon, Enhance, JointGraph}.

In this paper, we propose a simple yet effective unsupervised object localization method based on the representer theorem using self-supervised representation learning models.
The proposed method formulates the solution to the object localization task based on representer point selection~\cite{RPS}.
This strategy is derived from the representer theorem, where the predictions of a neural network for a given test point find expression through a linear combination of the activations originating from training examples.
The importance of each feature, which measures the significance of training examples in contributing to the final prediction, is computed by its norm. 
By selecting representer points that have the most influence on the model's predictions, our model gains valuable insights into how the model identifies foreground regions.
In addition, we introduce a simple extension to weakly supervised object localization, demonstrating zero-shot transferability across different datasets without the need for additional training.

Our experiments are conducted on various datasets for the object localization task to prove the effectiveness of our method and show outstanding performances compared to the state-of-the-art methods with substantial margins.
Furthermore, as shown in Figure~\ref{fig:graph}, our method even outperforms some recent weakly supervised and few-shot object localization methods without any fine-tuning or post-processing using additional refinement networks.

We summarize our contributions as follows:
\begin{itemize}
    \item We propose a novel unsupervised object localization method that leverages the representer theorem on self-supervised representation learning models. The proposed method helps to enhance the interpretability of predictions by providing relevant examples as well as their importance.
    
    \item We show that the representer point selection, which has not been extensively explored in previous research on deep neural networks, can be effectively applied to object localization.
    
    \item Our method is simple to implement and works well with a variety of self-supervised visual representation learning models. It is also an effective solution due to its good transferability across datasets and easy scalability to weakly supervised object localization.
    
    \item  Our method shows outstanding results on various datasets for the object localization task and also achieves comparable performances to the state-of-the-art weakly-supervised and few-shot methods.
\end{itemize}

\section{Related Works}

\paragraph{Weakly Supervised Object Localization}
Weakly supervised object localization (WSOL) is a challenging task, aiming to localize an object with only image-level annotations.
CAM~\cite{CAM} is the first work to introduce a method of generating a class activation map by projecting the weights of the classifier on the feature maps.
However, CAM has a limitation in finding parts of the objects that are not relevant for classification. 
To address this problem, various CAM-based WSOL methods have been proposed~\cite{HaS, ACoL, ADL, SPG, cream, FAM, CCAM, BAS, fcam}.
HaS~\cite{HaS} used the random masking of image patches to encourage the model to observe more object regions.
ACoL~\cite{ACoL} adopted two adversarial classifiers to catch the relevant parts.
{ADL~\cite{ADL} and AE~\cite{AE} progressively erased the discriminative parts on the feature map of each convolutional layer, ensuring that the focus was not solely on these discriminative areas.
SPG~\cite{SPG} and I$^2$C~\cite{I2C} utilized shallow convolutional feature maps and pixel-level correlations between two different images to adjust an activation map respectively.
PSOL~\cite{PSOL} proposed a new task of pseudo-supervised object localization which consists of two independent tasks: class-agnostic object localization and object classification.
PSOL~\cite{PSOL} and SEM~\cite{SEM} trained a localizer with an object bounding box generated by DDT~\cite{DDT} and edges of an input image extracted from the canny edge detector as pseudo labels, respectively.
{Recently, most CAM-based methods~\cite{cream, SPOL, FAM, CCAM, BAS, fcam} have focused on learning discriminative features for the foreground and background, while vision transformer-based methods~\cite{TSCAM, vitol, lctr, SCM} have emerged to exploit long-range dependency.}

\vspace{-0.2cm}
\paragraph{Unsupervised Object Localization} 
Unsupervised object discovery is strongly related to co-localization~\cite{DDT, Psy, Cocon} and co-segmentation~\cite{cyclesegnet, deepcoseg} tasks.
Early work for unsupervised object discovery~\cite{partbased} proposed a part-based region matching using off-the-shelf region proposals.
With the advance of deep learning, DDT~\cite{DDT} and SCDA~\cite{SCDA} introduced methods to mine the statistical properties of object regions in an image using features from pre-trained CNNs.
MO~\cite{OLM} mined a frequently activated pattern from multi-scale feature maps.
These methods achieved {impressive} results not only in object discovery but also in co-localization, which aims to localize objects in a dataset that consists of only one coarse category.
PsyNet~\cite{Psy} introduced a simple and practical framework for the co-localization task using self-supervised learning and class-agnostic activation mapping.
Ki~\etal~\cite{Cocon} employed contrastive learning with the assumption of consistency between image-level and activation-level transformations.
Su~\etal~\cite{JointGraph} proposed a joint graph partition method splitting foreground and background regions in a paired image feature graph.
C$^2$AM~\cite{CCAM} proposed to apply contrastive learning to disentangle the feature map into foreground and background with an additional trainable layer generating a class-agnostic activation map.
LOST~\cite{LOST} and TokenCut~\cite{tokencut} utilize a pre-trained vision transformer model~\cite{VIT} by DINO~\cite{dino}, which has the ability to produce fine segmentation masks using attention maps with {[CLS]} token.
LOST~\cite{LOST} suggested a heuristic seed selection and expansion method, and TokenCut~\cite{tokencut} employed {a normalized cut algorithm on the graph using the similarity between patches.}

\vspace{-0.2cm}
\paragraph{Self-supervised Visual Representation Learning}
Self-supervised visual representation learning (SSL) methods have been proposed to learn meaningful visual representation without manual annotations.
The network learns visual representation by training pretext tasks rather than the task intended to be solved.
For example, Noroozi~\etal~\cite{jigsaw} trained CNN to solve a jigsaw puzzle, Larsson~\etal~\cite{colorization} reconstructed color images from gray-scale images, and Gidaris~\etal~\cite{RotNet} predicted discrete rotation angles of rotated images.
Recently, contrastive learning-based SSL methods~\cite{mocov1, simclr, simsiam, BYOL} have emerged with comparable performances to the supervised learning methods.
They focus on modeling image similarity and dissimilarity between pairs of images~\cite{simclr,mocov1} or only consider image similarity~\cite{BYOL,simsiam} based on data augmentation.
DINO~\cite{dino} proposed a knowledge distillation strategy to train the vision transformer model.
These pre-trained SSL models are utilized in various downstream tasks of computer vision with fine-tuning.
On the contrary, in this work, we focus on leveraging the pre-trained SSL models for the unsupervised object localization task without any fine-tuning.

\section{Background}
\label{sec:background}
\paragraph{Representer Theorem.} 
Representer theorem~\cite{RT} provides a mathematical basis for traditional machine learning methods.
Let $L(\mathbf{x}_n, y_n, \mathbf{w})$ be the loss function, and $\frac{1}{N}\sum_{n=1}^N L(\mathbf{x}_n, y_n, \mathbf{w})$ be the empirical risk over a reproducing kernel Hilbert space (RKHS) $H_k$.
Then the representer theorem states that an optimal solution of a regularized empirical risk minimization (ERM) problem can be represented as a linear combination of a positive definite kernel $k$ on the input set $\mathcal{X}$ over $H_k$. 
The optimal parameter $\mathbf{w}^\ast$ can be expressed through ERM as follows:
\begin{align}
    \mathbf{w}^\ast=\underset{\mathbf{w}\in H_k}{\mathrm{argmin}} \{\frac{1}{N}\sum_{n=1}^N L(\mathbf{x}_n, y_n, \mathbf{w}) + \lambda ||\mathbf{w}||^2\}.
\label{eq:erm}
\end{align}
can be rewritten by the representer theorem
\begin{equation}
\begin{aligned}
    &\mathbf{w}^\ast (\cdot) = \sum_{i=1}^{n} \alpha_{i} k(\cdot,\mathbf{x}_i)=\sum_{i=1}^{n} \alpha_{i} \langle \varphi(\cdot) , \varphi(\mathbf{x}_i) \rangle,
\end{aligned}
\end{equation}
where $\alpha_i \in\mathbb{R}$, and $\varphi$ denotes a mapping function into {$H_k$}.
For further details for the Representer theorem, please refer to the work by \cite{RT, RPS}.

\vspace{-0.2cm}
\paragraph{Representer Point Selection for DNNs.}
The representer theorem is originally developed for non-parametric predictors, where the parameters lie in a reproducing kernel Hilbert space.
In deep neural networks, however, it is difficult to find a global solution for empirical risk minimization. 
To address this issue, \cite{RPS} proposes a technique to decompose the pre-activation predictions into a linear combination of the activation values of the training points. 
The neural network can be represented as
\begin{align}
{\hat{y}}_t=\sigma(\phi(\Phi(\mathbf{x}_t, \mathbf{W}_{\Phi}),\mathbf{w}_{\phi}))=\sigma(\mathbf{w}_\phi ^\top \mathbf{f}_t),
\end{align}
where $\sigma$ is the activation function, $\mathbf{f}_n=\Phi(\mathbf{x}_n, \mathbf{W}_{\Phi})$ is the last intermediate layer feature of a training example $\mathbf{x}_n$ in DNNs, $\phi$ denotes the last prediction layer for a specific task, and $\mathbf{W}_\Phi$ and $\mathbf{w}_\phi$ denote the parameters of $\Phi$ and $\phi$, respectively.
Let $\mathbf{W}^*$ be a stationary point of the objective \eqref{eq:erm}.
Then we have decomposition as follows:
\begin{align}
    {\mathbf{w}^*}_\phi^\top \mathbf{f}_t = \sum_{n=1}^N \alpha_n  k(\mathbf{x}_t, \mathbf{x}_n) = \sum_{n=1}^N\alpha_n {\mathbf{f}^\top_n} \mathbf{f}_t,\\
    \text{where~~} \alpha_n=\frac{1}{-2\lambda N} \frac{\partial L(\mathbf{x}_n,{y}_n,\mathbf{W})}{\partial \phi(\mathbf{f}_n,\mathbf{w}_\phi)},
\end{align}
and $\alpha_n {\mathbf{f}^\top_n} \mathbf{f}_t$ is a representer value of a training sample $\mathbf{x}_n$ given a test sample $\mathbf{x}_t$.
This representer value is assigned to each training point reflect their importance on the learned parameters and $\alpha_n$ is the global sample importance which is used to evaluate the importance of the training data $\mathbf{x}_n$.
Consequently, this approach enables the selection of representer points -- training instances with either large or small representer values -- to enhance comprehension of the model's predictions.

\begin{figure*}
    \centering
    \includegraphics[width=1\linewidth]{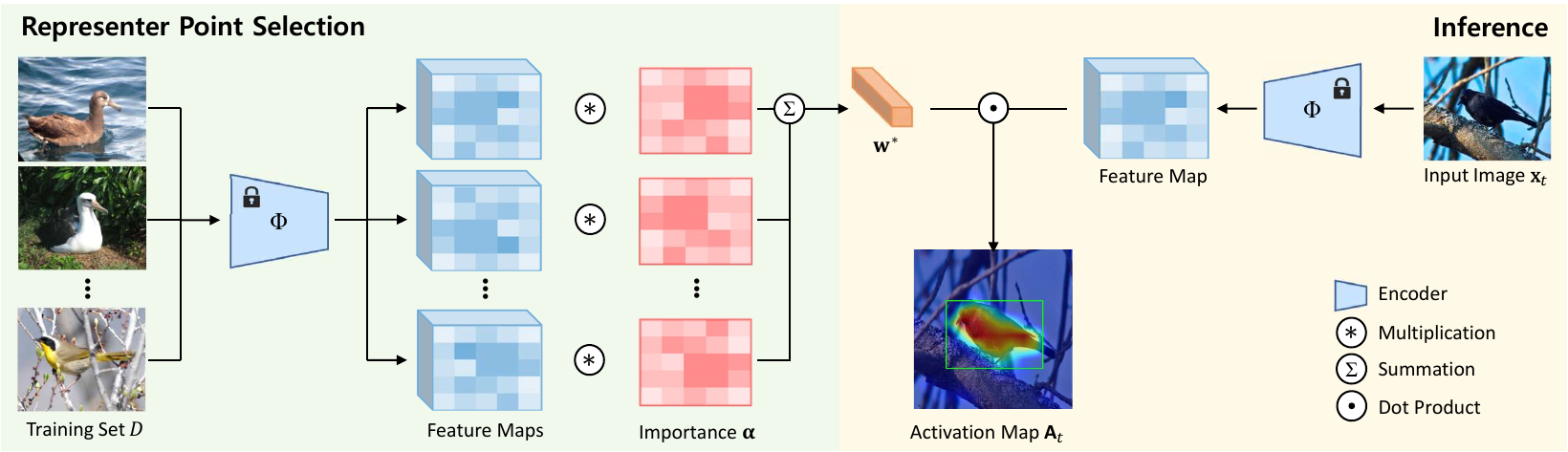}
    \caption{Pipelines of our unsupervised object localization method, which consists of two separate stages: representer point selection and inference.
    In the representer point selection stage, we compute a foreground predictor $\mathbf{w}^*$ using images in the training dataset.
    In the inference stage, the activation maps are computed by a dot product between the feature map of the test image and $\mathbf{w}^*$.
    }
    \label{fig:framework}
    \vspace{-0.2cm}
\end{figure*}

\section{Method}
\label{sec:method}
In this section, we discuss how we leverage the representer point selection~\cite{RPS} to solve unsupervised object localization (UOL) task, and our extension to weakly supervised object localization (WSOL) and zero-shot transferring.

\subsection{Representer Point Selection for UOL}
\label{subsec:apply}

Inspired by the representer point selection, we formulate the object localization task as identifying the important regions in training examples, called representer points, using a pre-trained self-supervised model.
Unlike other tasks such as classification, regression, and retrieval that are covered in \cite{RPS}, the object localization task requires dense predictions within a single image.
To address this, we address all elements within image feature maps rather than relying solely on a single image-level feature vector.

As our focus lies in unsupervised learning, the empirical risk or the loss function can be defined as $\frac{1}{N_D}\sum_{n=1}^{N_D} L(\mathbf{x}_n, \mathbf{W}_{\Phi})$ without a label $y_n$ given pre-trained encoder $\Phi$ and training set $D$.
Then the activation maps $\mathbf{A}_t$ for foreground regions can be computed by aggregating feature maps using the global sample importance, which indicate how much each training example contributes to the prediction of being foreground:
\begin{align}
    {\text{vec}(\mathbf{A}_{t})_{j}}= \underbrace{\sum_{n=1}^{N_D}\sum_{i=1}^{WH} \alpha_{n,i} {\mathbf{\hat{f}}_{n,i}^{\top}} }_{\mathbf{w}^*} \mathbf{\hat{f}}_{t,j} = {\mathbf{w}^*}^\top \mathbf{\hat{f}}_{t,j},
\label{eq:act}
\end{align}
where $\mathbf{f}_{n,i}=\Phi(\mathbf{x}_{n}, \mathbf{W}_\Phi)_{i}$ denotes $i^{\text{th}}$ feature vector in the feature map of an image $\mathbf{x}_{n}$,  which has been encoded using a self-supervised pre-trained encoder $\Phi$, $\mathbf{\hat{f}}=\frac{\mathbf{f}}{\|\mathbf{f}\|}$ indicates the normalized feature vector, $\alpha$ is the global sample importance, and $N_D$, $W$, $H$ denote the number of images in the training dataset, the width of the feature map, and the height of the feature map, respectively.
Here, we refer to $\alpha_{n,i} {\mathbf{\hat{f}}_{n,i}^{\top}} \mathbf{\hat{f}}_{t,j}$ as the \textit{representer value} for each training image patch $\mathbf{x}_{n,i}$ given a testing image patch $\mathbf{x}_{t,j}$, corresponding to receptive fields of $\mathbf{f}_{n,i}$ and $\mathbf{f}_{t,j}$, respectively. 
In Figure~\ref{fig:framework}, we illustrate the entire structure of the proposed representer point selection method to compute a foreground predictor $\mathbf{w}^*$ and inference pipelines using it.
\vspace{-0.2cm}

\begin{figure*}
    \centering
    \includegraphics[width=0.95\linewidth]{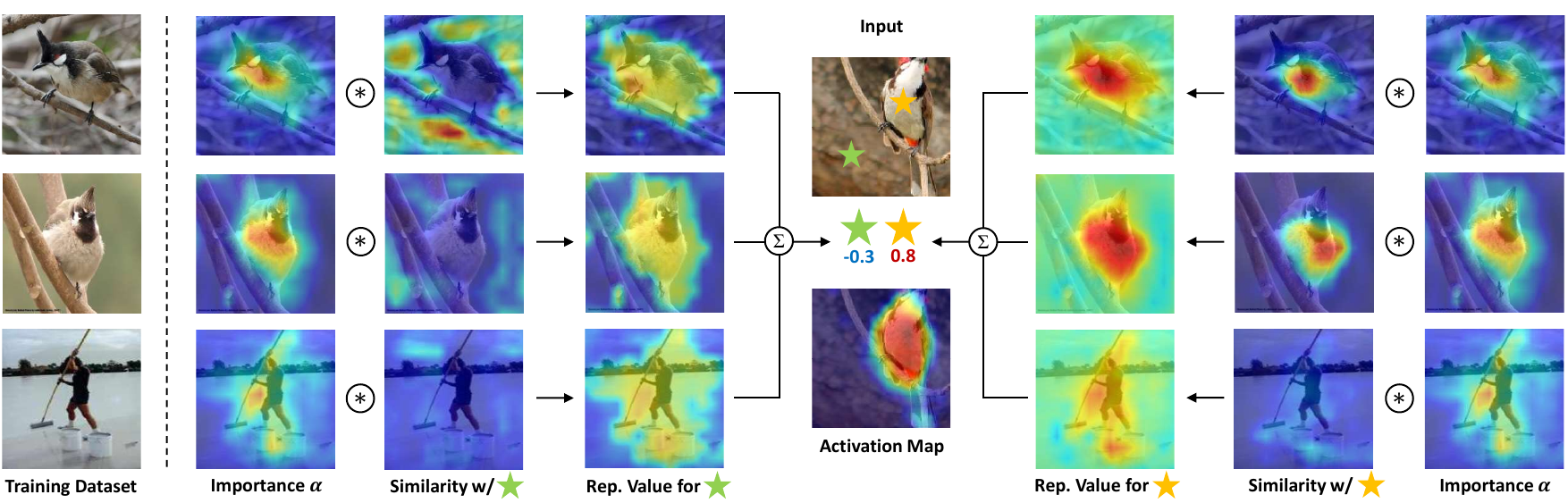}
    \caption{Illustration of how our method computes activation maps using two example points, yellow and green stars. To predict the point as a foreground region, two factors should be considered: the global sample importance and the similarity of features. the representer values are obtained by multiplying these two factors, and a high representer value indicates a stronger contribution to the foreground prediction.
    }
    \label{fig:simmapf}
    \vspace{-0.3cm}
\end{figure*}

\paragraph{Global Sample Importance.}
\label{subsec:alpha}
$\alpha_n$ is referred to as the global sample importance since it is independent of the testing examples but reliant on the empirical risk of the training set. 
In order to tackle the unsupervised localization task, we first design the loss function which is suitable for an unsupervised setting to compute $\alpha$.
Let $\mathbf{w}$ be the parameters of a binary classifier for foreground predictions given feature vectors, then the loss function can be written as follows:
\vspace{-0.2cm}
\begin{align}
    L(\mathbf{x}_{n,i},{y}_{n,i},\mathbf{W}_{\Phi}) = - y_{n,i} \mathbf{w}^{\top} {\mathbf{\hat{f}}_{n,i}},
\end{align}
where ${y}_{n,i}\in\{-1,1\}$ denotes a label of $\mathbf{x}_{n,i}$, which is an image patch corresponding to a receptive field of $\mathbf{f}_{n,i}$.
In the unsupervised object localization task, however, the goal is to find foreground regions in images without the use of labels.
Thus, instead of using $y_{n,i}$, we use the norm of the feature vector to determine a soft pseudo label of the example for the objective function as follows:
\begin{align}
    L(\mathbf{x}_{n,i},\mathbf{W}_{\Phi}) = - (\|\mathbf{f}_{n,i}\| - \tau) \mathbf{w}^{\top} {\mathbf{\hat{f}}_{n,i}},
\end{align}
where $\tau$ denotes a foreground threshold.

We estimate soft labels of inputs based on the norm of feature vectors because the norm can serve as an important indicator of the features that the network emphasized during training with the self-supervised contrastive loss. 
In other words, a higher norm of a feature vector indicates that the corresponding feature had more weight in the model training.
These features are extracted from regions of the image that contain rich information to distinguish them from other instances, and these regions are more likely to be associated with significant objects.
Likewise, prior self-supervised co-localization studies~\cite{Psy, Cocon, Enhance, JointGraph} also use the magnitude of feature vectors as a clue to discover object regions.


Hence, the global sample importance is computed as:
\vspace{-0.4cm}
\begin{align}
    \alpha_{n,i}&=-\frac{1}{2\lambda N_DHW}\frac{\partial L(\mathbf{x}_{n,i},\mathbf{W}_{\Phi})}{\partial \mathbf{w}^\top \mathbf{\hat{f}}_{n,i}}\\
    &= \frac{1}{2\lambda N_DHW}(||\mathbf{f}_{n,i}||-\tau)=\frac{1}{C}(||\mathbf{f}_{n,i}||-\tau).
\label{eq:alpha}
\end{align}
Here, the sign of the global sample importance is related to whether to \textit{representer points} -- training examples that have large or small representer values -- contribute to the foreground or background.
In order to have large representer values, both $\alpha$ and similarity between two features, expressed as $\mathbf{f}_n^\top\mathbf{f}_t$, should have large values.
By analyzing the representer values, we can gain insight into how the model predicts foreground regions by providing relevant examples as well as their importance.
Figure~\ref{fig:simmapf} shows an example of global sample importance, the similarity between features, and representer values for given points in a given test image.
\vspace{-0.5cm}

\paragraph{Straightforward Computation of $\mathbf{w}^*$}
We then compute a foreground predictor $\mathbf{w}^*$ using $\alpha_{n,i}$ in Eq~\eqref{eq:alpha} as follows:
\vspace{-0.1cm}
\begin{align}
    \mathbf{w}^{*}&= \frac{1}{C}\underbrace{\sum_{n=1}^{N_D}\sum_{i=1}^{HW} \mathbf{f}_{n,i}}_{\mathbf{v}} - \frac{\tau}{C}\underbrace{\sum_{n=1}^{N_D}\sum_{i=1}^{HW} \mathbf{\hat{f}}_{n,i}}_{\mathbf{u}},
    \vspace{-0.2cm}
\end{align}
The optimal $\mathbf{w}^*$ can be computed by taking the difference between the mean feature vector of all images in the dataset and the mean of the normalized feature vectors of all images in the dataset, multiplied by two constant values.
Algorithm~\ref{algo:your-algo} illustrates the Python-style pseudo code of the proposed representer point selection method to compute the foreground predictor $\mathbf{w}^*$ and it clearly shows the simplicity and reproducibility of our method.

\begin{algorithm}[t]
\small
\SetAlgoLined
    \PyComment{model: pre-trained encoder $\Phi$} \\
    \PyComment{const: the constant value for $\alpha$}\\
    \PyComment{th: foreground threshold of norm $\tau$}\\
    \PyCode{for x in data loader:} \\
    \PyCode{\quad f = model(x)} \PyComment{f:[N,C,H,W]}\\ 
    \PyCode{\quad v += sum(f,d=(N,H,W))} \\
    \PyCode{\quad u += sum(normalize(f,d=C),d=(N,H,W))} \\
    \PyCode{w = v/const - th*u/const} \\
    \PyCode{return w} \\
\caption{Representer Point Selection}
\label{algo:your-algo}
\end{algorithm}


\subsection{Towards WSOL and Zero-shot Transferring}
\label{subsec:wsol}
Our method can be easily extended to weakly supervised and zero-shot transferring settings with a simple modification, since activation maps are computed by the linear combination of activations of training points directly without learning.
Given a class label $c$, the activation map $\mathbf{A}^c$ is computed as follows:
\vspace{-0.2cm}
\begin{align}
    {\text{vec}(\mathbf{A}_{t}^c)_{j}} = \sum_{n\in{D_c}}\sum_{i=1}^{WH} \alpha_{n,i} {\mathbf{\hat{f}}_{n,i}^{\top}} \mathbf{\hat{f}}_{t,j} = {\mathbf{w}_c^*}^\top \mathbf{\hat{f}}_{t,j},
\end{align}
where $D_c$ denotes the subset of the training dataset $D$ that contains only the images assigned to class $c$.
Note that, although WSOL allows for the use of image-level class labels, we utilize an off-the-shelf classification model to determine the class of each image in $D$ during our experiments in order to ensure a fair comparison with other methods~\cite{CCAM, PSOL}.
Extending our method to zero-shot transferring scenarios can be simply done by computing the weight vector $\mathbf{w^*}$ on the different datasets in which the classes do not overlap with the test dataset.


\section{Experiments}
\label{sec:experiments}

\begin{table*}
\footnotesize
  \centering
    \caption{Comparison between our method and the state-of-the-art self-supervised/unsupervised methods in terms of \textit{GT-Known Loc} performance on ImageNet-1K and four fine-grained datasets: CUB200-2011, Stanford Cars, FGVC-Aircraft, and Stanford Dogs. $\ast$ denotes our reproduced results from their official code. $\dagger$ and $\ddagger$ indicate a backbone initialized by MoCo~v2 and DINO pre-trained weights, respectively.\vspace{0.1cm}}
  \begin{tabular}{lccccccc}
    \toprule
    Method & Backbone & $\mathcal{T}$ & CUB-200-2011 & Stanford Cars & FGVC-Aircraft & Stanford Dogs & ImageNet-1K \\
    \midrule
    \textit{Self-supervised methods} \\
    ORE $_{\text{'}22}$~\cite{Enhance} & VGG16 & \cmark & 87.72 & 97.43 & 97.86 & 85.45 & - \\
    JGP $_{\text{'}22}$~\cite{JointGraph} & VGG16 & \cmark & 88.83 & 97.73 & 96.72 & - & - \\
    PsyNet $_{\text{'}20}$~\cite{Psy} & SE-ResNet50 & \cmark & 85.10 & 98.81 & 97.81 & 77.84 & - \\
    Ki~\etal $_{\text{'}21}$~\cite{Cocon} & SE-ResNet50 & \cmark & 85.93 & 98.95 & 98.75 & 80.32 & - \\
    PSOL $_{\text{'}20}$~\cite{PSOL} & ResNet50 & \cmark & 90.00 & - & - & - & 65.44 \\
    C$^2$AM $_{\text{'}22}$~\cite{CCAM} & ResNet50$^\dagger$ & \cmark & 89.90 & 99.44$^\ast$ & 98.65$^\ast$ & 89.91$^\ast$ & 66.51 \\
    C$^2$AM (+PSOL) $_{\text{'}22}$~\cite{CCAM} & ResNet50$^\dagger$ & \cmark & 91.54 & - & - & - & 68.07 \\
    \midrule
    \textit{w/o finetuning} \\
    DDT $_{\text{'}19}$~\cite{DDT} & VGG16 & \xmark & 82.26 & 71.33 & 92.53 & - & - \\
    MO $_{\text{'}20}$~\cite{OLM} & VGG16 & \xmark & 80.45 & 92.51 & 94.94 & 80.70 & - \\
    LOST $_{\text{'}21}$~\cite{LOST} & ViT-S$^\ddagger$ & \xmark & 89.70 & - & - & - & 60.00 \\
    TokenCut $_{\text{'}22}$~\cite{tokencut} & ViT-S$^\ddagger$ & \xmark & 91.80 & - & - & - & 65.40 \\
    \hdashline[1pt/1pt]
    Ours & VGG16 & \xmark & 90.47 & 98.31 & 96.88 & 89.67 & 59.07 \\
    Ours & ResNet50$^\dagger$ & \xmark & \textbf{96.67} & 99.69 & 98.71 & \textbf{95.07} & 66.89 \\
    Ours & ViT-S$^\ddagger$ & \xmark & 91.61 & \textbf{99.84} & \textbf{99.22} & 94.71 & \textbf{73.65} \\
    \bottomrule
  \end{tabular}
  \vspace{-0.2cm}
  \label{tab:CoLoctable}
\end{table*}


\begin{table*}
\footnotesize
  \centering
  \vspace{0.3cm}
    \caption{Comparison of localization performances in terms of \textit{Top-1}, \textit{Top-5} and \textit{GT-known Loc} on CUB-200-2011 test set and ImageNet-1K validation set. 
    Self-supervised and unsupervised methods including ours employ an additional classifier to evaluate in WSOL setup following PSOL~\cite{PSOL}.
    $\mathcal{T}$ and $\mathcal{S}$ indicate whether each method uses training and image-level supervision (B for bounding box annotations), respectively. $\dagger$, $\ddagger$, and $\circ$ indicate the use of MoCo~v2, BYOL, and DINO pre-trained weights, respectively.\vspace{0.1cm}}
  \begin{tabular}{lcccccccccc}
    \toprule
    \multirow{2}{*}{Method}& \multirow{2}{*}{Backbone} & \multirow{2}{*}{$\mathcal{S}$} & \multirow{2}{*}{$\mathcal{T}$} & \multicolumn{3}{c}{CUB-200-2011} & & \multicolumn{3}{c}{ImageNet-1K} \\ \cmidrule{5-7} \cmidrule{9-11}
    & & & & \textit{Top-1 Loc} & \textit{Top-5 Loc} & \textit{GT-Known} && \textit{Top-1 Loc} & \textit{Top-5 Loc} & \textit{GT-Known} \\
    \midrule
    \textit{Few-shot method}\\
    SPNet $_{\text{'}19}$~\cite{spnet} & ResNet50 & \cmark (B)  & \cmark & - & - & 95.80 & & - & - & 71.50 \\
    \midrule
    \textit{Weakly supervised methods}\\
    FAM $_{\text{'}21}$~\cite{FAM} & ResNet50 & \cmark  & \cmark & 73.74 & - & 85.73 & & 54.46 & - & 64.56 \\
    CREAM $_{\text{'}22}$~\cite{cream} & ResNet50 & \cmark & \cmark & 76.03 & - & 89.88 & & 55.66 & - & 69.31 \\
    BGC $_{\text{'}22}$~\cite{bgc} & ResNet50 & \cmark  & \cmark & 73.16 & 86.68 & 91.60 & & 53.76 & 65.75 & 69.89 \\
    DAOL $_{\text{'}22}$~\cite{DAWSOL} & ResNet50 & \cmark  & \cmark & 66.65 & - & 81.83 & & 55.84 & - & 70.27 \\
    BAS $_{\text{'}22}$~\cite{BAS} & ResNet50 & \cmark  & \cmark & 77.25 & 90.08 & 95.13 & & 57.18 & 68.44 & 71.77 \\
    BagCAM$_{\text{'}22}$~\cite{Bag}  & ResNet50 & \cmark  & \cmark & 69.67 & - & 94.01 & & 44.24 & - & 72.08 \\
    TS-CAM $_{\text{'}21}$~\cite{TSCAM} & ViT-S & \cmark  & \cmark  & 71.30  & 83.80 & 87.70 & & 53.40 & 64.30 & 67.60 \\
    LCTR $_{\text{'}22}$~\cite{lctr} & ViT-S & \cmark  & \cmark  & 79.20 & 89.90 & 92.40 & & 56.10 & 65.80 & 68.70 \\
    ViTOL $_{\text{'}22}$~\cite{vitol} & ViT-S & \cmark  & \cmark  & -  & - & 80.90 & & 58.64 & - & 72.51 \\
    SCM $_{\text{'}22}$~\cite{SCM} & ViT-S & \cmark  & \cmark  & 76.40  & 91.60 & 96.60 & & 56.10 & 66.40 & 68.80 \\
    \midrule
    \textit{Self-supervised methods}\\
    PSOL $_{\text{'}20}$~\cite{PSOL} & ResNet50 & \xmark  & \cmark & 70.68 & 86.64 & 90.00 & & 53.98 & 63.08 & 65.44 \\
    C$^2$AM $_{\text{'}22}$~\cite{CCAM} & ResNet50$^\dagger$ & \xmark  & \cmark & -  & - & 89.90 & & - & - & 66.51 \\
    C$^2$AM~(+PSOL) $_{\text{'}22}$~\cite{CCAM} & ResNet50$^\dagger$ & \xmark  & \cmark & 74.76 & 87.37 & 91.54 & & 54.65 & 65.05 & 68.07 \\
    \midrule
    \textit{w/o finetuning}\\
    LOST $_{\text{'}21}$~\cite{LOST} & ViT-S$^\circ$ & \xmark  & \xmark & 71.30 & - & 89.70 & & 49.00 & - & 60.00 \\
    TokenCut $_{\text{'}22}$~\cite{tokencut} & ViT-S$^{\circ}$ & \xmark  & \xmark & 72.90 & - & 91.80 & & 52.30 & - & 65.40 \\
    \hdashline[1pt/1pt]
    Ours & ResNet50$^{\dagger}$ & \xmark  & \xmark & \textbf{79.57} & \textbf{92.60} & 96.67 & & 55.60 & 66.05 & 69.10 \\
    Ours & ResNet50$^{\ddagger}$ & \xmark  & \xmark & 79.55 & 92.58 & \textbf{96.72} & & 56.03 & 66.78 & 69.93 \\
    Ours & ViT-S$^{\circ}$ & \xmark  & \xmark & 74.97 & 84.24 & 91.61 & &  {\textbf{62.03}} & {\textbf{71.96}} & \textbf{74.44} \\
    \bottomrule
  \end{tabular}
  \vspace{-0.2cm}
  \label{tab:WSOLtable}
\end{table*}

\begin{table}
\footnotesize
  \centering
    \caption{Comparison between the proposed method and the state-of-the-art weakly supervised object localization methods in terms of \textit{PxAP} and \textit{PIoU} on CUB-200-2011 and OpenImages-30K. $\ast$ denotes that the results are obtained from their official code. \vspace{0.1cm}}
  \begin{tabular}{lccccc}
    \toprule
    \multirow{2}{*}{Method} & \multicolumn{2}{c}{CUB-200-2011} & & \multicolumn{2}{c}{OpenImages-30K} \\ \cmidrule{2-3} \cmidrule{5-6}
    & \textit{PIoU} & \textit{PxAP} & & \textit{PIoU} & \textit{PxAP}\\
    \midrule
    \multicolumn{5}{l}{\textit{WSOL methods}} \\
    BGC $_{\text{'}22}$~\cite{bgc} & - & - & & - & 63.70 \\
    CREAM $_{\text{'}22}$~\cite{cream} & - & - & & - & 64.70 \\
    DAOL $_{\text{'}22}$~\cite{DAWSOL} & 56.18 & 74.70 & & 49.68 & 65.42 \\
    BagCAM$_{\text{'}22}$~\cite{Bag} & 74.51 & 90.38 & & 52.17 & 67.68 \\
    \hline
    \multicolumn{5}{l}{\textit{Self-supervised methods}} \\  
    C$^2$AM $_{\text{'}22}$~\cite{CCAM} & 69.65$^\ast$ & 88.74$^\ast$ & & 47.75$^\ast$ & 58.28$^\ast$ \\
    \hline
    \multicolumn{5}{l}{\textit{Ours with SSL/WSOL pre-trained backbone}} \\  
    MoCo~v2~\cite{mocov2} + Ours & 71.28 & 87.03 & & 50.20 & 62.18\\
    DenseCL~\cite{densecl} + Ours & 70.11 & 87.74 & & 54.68 & {68.32}\\
    \textcolor{gray}{DINO~ViT-S~\cite{dino} + Ours} & \textcolor{gray}{78.97} & \textcolor{gray}{94.37} & & \textcolor{gray}{61.24} & \textcolor{gray}{73.14}\\
    \hdashline[1pt/1pt]
    C$^2$AM~\cite{CCAM} + Ours & 71.54 & 90.22 & & 56.15 & 69.79\\
    DAOL~\cite{DAWSOL} + Ours & 67.37 & 85.99 & & 53.25 & 67.62\\
    \bottomrule
  \end{tabular}
  \label{tab:pxap}
\end{table}

\subsection{Implementation Details}
\label{subsec:implementation}
We adopt ResNet50~\cite{resnet} and ViT-S~\cite{VIT} as backbones for SSL pre-trained model, utilizing publicly available pre-trained weights.
We use the input image size of 224$\times$224 for the representer point selection.
To evaluate object localization performances on fine-grained classification datasets, the input image is resized to 480$\times$480 and then center-cropped to a size of 448$\times$448.
In the case of ImageNet-1K, the input image is resized to 256$\times$256 and then center-cropped to a size of 224$\times$224. 
We use ${\|\mathbf{v}\|}/{\|\mathbf{u}\|}$ for $\tau$, and min-max normalization to estimate the bounding box from the generated activation map.
We also set $\lambda=0.001$ as in \cite{RPS}, but changing this value does not have an impact on the final results due to the use of min-max normalization.

\subsection{Datasets and Evaluation Metrics}
We conduct experiments on the four fine-grained datasets for unsupervised object localization to evaluate our method: CUB-200-2011~\cite{CUB}, Stanford Cars~\cite{CARS}, FGVC-Aircraft~\cite{AIRCRAFT}, and Stanford Dogs~\cite{DOGS}.
We also evaluate our method on ImageNet-1K to compare with other state-of-the-art methods in unsupervised and WSOL setups.
In addition, we evaluate the segmentation quality of our method by comparing it with the WSOL state-of-the-art methods on CUB-200-2011 and OpenImages-30K~\cite{openimages} datasets. 

Following~\cite{CAM}, we use Top-1 (\textit{Top-1 Loc}), Top-5  (\textit{Top-5 Loc}) and GT-known localization (\textit{GT-known Loc}) accuracy for evaluation.
\textit{GT-known Loc} computes the ratio of the images where the Intersection over Union (IoU) between the estimated bounding box and the known ground-truth boxes is greater than 50\%.
\textit{Top-1 Loc} and \textit{Top-5 Loc} compute the ratio of the images correctly classified with Top-1 and Top-5 predictions, respectively, with the condition of \textit{GT-Known Loc}.
For segmentation evaluation, \textit{Pixel-wise average precision~(PxAP)} computes the area under the pixel precision-recall curve and 
\textit{Peak-IoU (PIoU)} is the best IoU score with various thresholds.

\subsection{Results}
\paragraph{Unsupervised Object Localization.}
Our method is primarily designed for the unsupervised object localization (UOL) task which does not focus on localizing objects for specific categories. 
In this task, $\mathbf{w}^*$ is computed using the entire dataset, regardless of the number of categories in the dataset. 
We report the performances of our method and the state-of-the-art methods in terms of \textit{GT-known Loc} on five commonly used datasets in Table~\ref{tab:CoLoctable}.
We compare our method with recent self-supervised object localization methods including ORE~\cite{Enhance}, PsyNet~\cite{Psy}, Ki \etal~\cite{Cocon}, JGP~\cite{JointGraph} and C$^2$AM~\cite{CCAM}, as well as object localization methods without finetuning including DDT~\cite{DDT}, MO~\cite{OLM}, LOST~\cite{LOST}, and TokeneCut~\cite{tokencut}.
As shown in Table~\ref{tab:CoLoctable}, our method significantly surpasses other methods on all benchmark datasets.
Our method outperforms not only unsupervised methods but also the self-supervised methods where the models are finetuned on the training split of the target datasets.
Even with a shallower architecture, such as VGG-16, our method shows better performance compared to the methods using ResNet50 on CUB-200-2011.
For the VGG-16 backbone, we employ ImageNet pre-trained weights.

\begin{table}
\footnotesize
  \centering
    \caption{\textit{GT-Known Loc} results of our method on various SSL models with ResNet50 backbones except DINO~ViT.\vspace{0.1cm}}
    \begin{tabular}{lcccc}
    \toprule
    Model & BYOL & MoCo v2 / v3 & DINO Res / ViT \hspace{-0.3cm} & SimSiam\\
    \midrule
    CUB & 96.72 &  96.67 / 97.03 & 96.15 / 91.61 & 94.99 \\
    ImageNet & 67.05 & 66.89 / 66.31 & 65.15 / 73.65 & 64.10 \\
    ImageNet ($\mathbf{w}_c^*$)\hspace{-0.3cm} & 69.93 & 69.10 / 69.34 & 69.15 / 74.44 & 67.57 \\
    \bottomrule
  \end{tabular}
  \label{tab:Othernet}
\end{table}

\begin{table}
\footnotesize
  \centering
    \caption{\textit{GT-Known Loc} results of models with and without our inference using the representer point selection.\vspace{0.1cm}}
  \begin{tabular}{lll}
    \toprule
    Method & CUB & ImageNet \\
    \midrule
    MoCo~v2 + CAAM & 86.93 & 61.19 \\
    MoCo~v2 + Ours & 96.67 (\textcolor{teal}{+9.74}) & 69.10  (\textcolor{teal}{+7.91}) \\
    \hdashline[1pt/1pt]
    C$^2$AM~\cite{CCAM} & 89.90 & 66.51\\
    C$^2$AM~\cite{CCAM} + PSOL & 91.54 (\textcolor{gray}{+1.64}) & 68.07 (\textcolor{gray}{+1.56}) \\
    C$^2$AM~\cite{CCAM} + Ours & 96.86 (\textcolor{teal}{+6.96}) & 69.80 (\textcolor{teal}{+3.29}) \\
    \hdashline[1pt/1pt]
    DAOL~\cite{DAWSOL} & 81.83 & 70.27 \\
    DAOL~\cite{DAWSOL} + Ours & 89.44 (\textcolor{teal}{+7.61}) & 70.97 (\textcolor{teal}{+0.70})\\
    \bottomrule 
  \end{tabular}
  \label{tab:unbiased}
\end{table}


\vspace{-0.2cm}
\paragraph{Weakly Supervised Object Localization.}

{In addition, as mentioned in Section~\ref{subsec:wsol}, our method can be extended to the weakly supervised object localization (WSOL) setting by computing $\mathbf{w}_c^*$ for each class $c$ using a pre-trained classifier.
To evaluate on \textit{Top-1/5} Loc metrics, we follow \cite{PSOL} which divides the WSOL task into two sub-tasks: object localization and classification. 
We first localize objects, and then we determine classes for localized objects using the pre-trained classifier, \ie ResNet50 pre-trained on ImageNet. 


Table~\ref{tab:WSOLtable} shows the performance of our method and the state-of-the-art weakly supervised object localization methods, including FAM~\cite{FAM}, CREAM~\cite{cream}, BGC~\cite{bgc}, BAS~\cite{BAS}, DAOL~\cite{DAWSOL}, BagCAM~\cite{Bag}, TS-CAM~\cite{TSCAM}, LCTR~\cite{lctr}, ViTOL~\cite{vitol}, and SCM~\cite{SCM} on CUB-200-2011 and ImageNet-1K datasets.
We also compare our method with PSOL~\cite{PSOL} and C$^2$AM~\cite{CCAM} which are self-supervised methods.
Our method shows outstanding results on CUB-200-2011 and comparable results on ImageNet-1K, compared to the advanced weakly supervised methods without any supervision and training.
As shown in Table~\ref{tab:WSOLtable}, our method outperforms all self-supervised methods with the additional classifier using the same architecture with localization backbone~\cite{resnet, VIT}.
For a fair comparison, we report C$^2$AM initialized with MoCo v2 pre-trained weights as well as its refinements with PSOL.
Note that, without any training and post-processing as PSOL, our method significantly surpasses C$^2$AM~(+PSOL) and PSOL on CUB-200-2011.
For a fine-grained dataset such as CUB-200-2011, we utilize a class-agnostic foreground predictor $\mathbf{w}^*$ for comprehensive object localization rather than a class-specific one.

Our method shows outstanding performance on not only localization but also segmentation, as shown in Table~\ref{tab:pxap}, particularly with the model trained by self-supervised pixel-wise contrastive learning~(\ie DenseCL~\cite{densecl}) on the OpenImages-30K dataset.
All the methods compared in this table employ ResNet50 as their backbone, with the exception of DINO.
Upon utilizing the DINO ViT model as our backbone, we observe a noteworthy and substantial enhancement in performance.
Exploiting our method as an inference method for feature extractors trained by the WSOL methods~\cite{CCAM, DAWSOL} further enhances performance, yielding significant improvements.
It clearly shows that our method demonstrates consistently good performances on various datasets and diverse metrics.

\begin{table}
\footnotesize
  \centering
    \caption{\textit{GT-Known Loc} results of C$^2$AM and ours using the model trained/computed on CIFAR$_{10}$, CIFAR$_{100}$, and CUB-200-2011 to show the zero-shot transferability of the methods.\vspace{0.1cm}}
  \begin{tabular}{lcccccc}
    \toprule
    Method & Training  & CUB & {Cars} & {Aircraft} & {Dogs} & \hspace{-0.2cm}ImgNet \\
    \midrule
    C$^2$AM & CIFAR$_{10}$ & 73.18 & 96.74 & 98.35 & 80.54 & 56.51 \\
    Ours & CIFAR$_{10}$ & 94.24 & 99.04 & 98.26 & 92.39 & 63.87 \\
    \hdashline[1pt/1pt]
    C$^2$AM & CIFAR$_{100}$ & 65.12 & 93.40 & 95.68 & 78.75 & 57.29 \\
    Ours & CIFAR$_{100}$ & 92.51 & 98.13 & 97.66 & 89.67 & 63.72 \\
    \hdashline[1pt/1pt]
    C$^2$AM & CUB & 89.90 & 90.85 & 89.80 & 90.44 & 59.08 \\
    Ours & CUB & 96.67 & 98.20 & 98.32 & 93.14 & 65.24 \\
    \bottomrule
  \end{tabular}
  \label{tab:transfer}
\end{table}

\begin{figure}[t]
    \centering
    \includegraphics[width=0.9\linewidth]{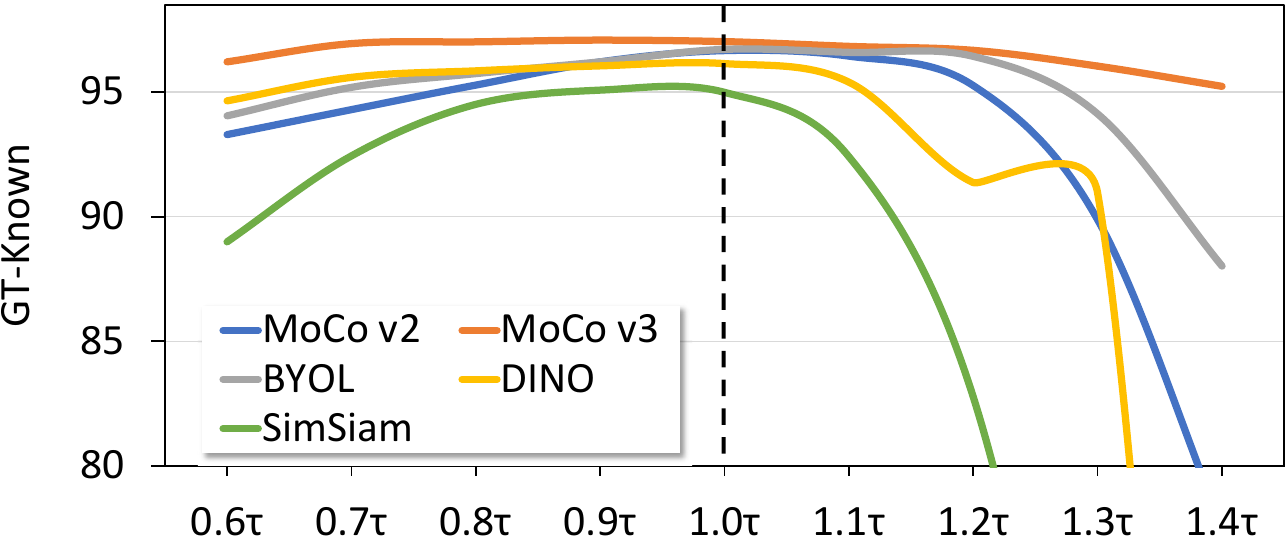}
    \caption{{\textit{GT-Known} scores with respect to different thresholds using various pre-trained models on the CUB-200-2011 dataset. The proposed threshold $\tau$ consistently achieves near-optimal scores.}}
    \label{fig:taugraph}
\end{figure}

\subsection{Ablation Study}
\label{subsec:ablation}
We conduct comprehensive ablation studies to verify the effectiveness of our method. 
We adopt the \textit{GT-Known Loc} metric as a performance evaluation for all experiments.

\vspace{-0.1cm}
\paragraph{Robustness across various SSL pre-trained models.}

We report \textit{GT-known Loc} scores of our method using popular self-supervised pre-trained models including BYOL~\cite{BYOL}, MoCo v2~\cite{mocov2}, MoCo v3~\cite{mocov3}, DINO~\cite{dino} and SimSiam~\cite{simsiam} to support compatibility of our method in Table~\ref{tab:Othernet}.
Our method consistently shows good performance with various pre-trained models.
{We believe that our method generates more precise activation map of the object, as long as good representations are provided by pre-trained models no matter how they are trained.}

\vspace{-0.1cm}

\paragraph{Effectiveness of representer point selection.}

Table~\ref{tab:unbiased} reports the results of our inference method using representer point selection on several models including C$^2$AM~\cite{CCAM} and DAOL~\cite{DAWSOL} as well as class-agnostic activation map (CAAM) proposed in~\cite{Psy}.
The backbones for our model and C$^2$AM are initialized by MoCo v2~\cite{mocov2} pre-trained weights.
Even with the state-of-the-art models trained for the object localization task, applying our inference improves their performance with significant margins on both CUB-200-2011 and ImageNet-1K datasets.
In the case of C$^2$AM, our inference significantly improves its performance more than using post-processing by PSOL~\cite{PSOL} which requires additional training using a ResNet50 network.

\begin{table}
\footnotesize
  \centering
    \caption{Ablation study on the dataset sampling ratio for the representer point selection. $\dagger$ denotes the use of a CIFAR$_{10}$ for the foreground predictor $\mathbf{w^*}$, and $\ddagger$ indicates the use of ResNet50 as the backbone; otherwise, ViT-s is employed.\vspace{0.1cm}}
  \begin{tabular}{lll}
    \toprule
    Method (sampling ratio) & CUB & ImageNet \\
    \midrule
    \textit{w/o finetuning} \\
    Ours (100\%) & 91.62 & 73.65 \\
    Ours (10\%) & 91.70 \scriptsize\textcolor{gray}{$\pm$0.11} & 73.64 \scriptsize\textcolor{gray}{$\pm$0.03}  \\
    Ours (1\%) & 91.65 \scriptsize\textcolor{gray}{$\pm$0.42} & 73.63 \scriptsize\textcolor{gray}{$\pm$0.03}  \\
    Ours (0.1\%) & 91.08 \scriptsize\textcolor{gray}{$\pm$1.02} & 73.67 \scriptsize\textcolor{gray}{$\pm$0.06} \\
    Ours (0\%$^\dagger$) & 92.73 & 73.10 \\
    \hdashline[1pt/1pt]
    TokenCut~\cite{tokencut} (100\%) & 91.80 & 65.40 \\
    \hline
    \textit{Weakly supervised method} \\
    ViTOL~\cite{vitol} (100\%) &  80.90 & 72.51\\
    \hline
    \textit{Few-shot method} \\
    SPNet$^\ddagger$~\cite{spnet} ($\sim$10\%/$\sim$1\%) &  95.80 & 71.50\\
    \bottomrule
  \end{tabular}
  \label{tab:sampling}
\end{table}

\begin{figure*}[t]
  \centering
    \includegraphics[width=0.95\linewidth]{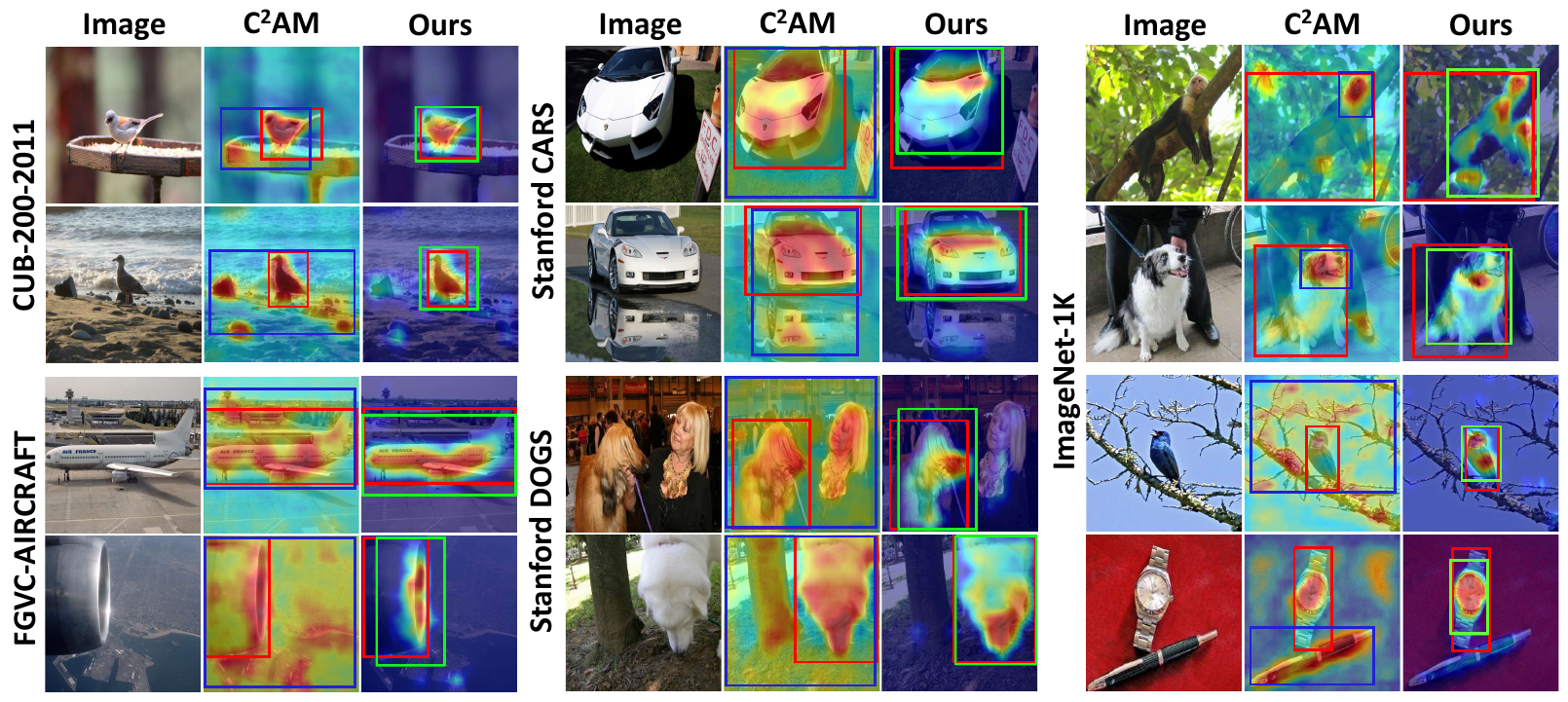}
    \caption{Qualitative results of our method compared to C$^2$AM. Ground truth bounding boxes are denoted in red, whereas predicted object localization bounding boxes are denoted in green or blue. Best viewed in color.}
    \label{fig:result}
\end{figure*}

\vspace{-0.1cm}
\paragraph{Zero-shot transferability across datasets.}

We also examine the transferability of our model across different datasets in Table~\ref{tab:transfer}.
We compare our method's performance with C$^2$AM~\cite{CCAM}, using our foreground predictor and C$^2$AM's trained model from CIFAR$_{10}$, CIFAR$_{100}$~\cite{CIFAR} and CUB-200-2011 datasets, and then testing on other datasets without additional training.
In this comparison, we ensure fairness by using class-agnostic weights and initializing the ResNet50 model with pre-trained weights from MoCo v2~\cite{mocov2}. 
The results in Table~\ref{tab:transfer} highlight that our model, leveraging representer points from these datasets, can be transferred to various datasets with strong performance, while C$^2$AM's performance notably drops.


\vspace{-0.1cm}
\paragraph{Impact of sampling rate on a dataset $D$.}

In order to show the impact of the number of samples for selecting representer points, we randomly sample images on the training set.
In Table~\ref{tab:sampling}, to illustrate the effect of sampling rate, we report the results of our method with DINO ViT-s using 100\%, 10\%, 1\%, 0.1\% and 0\% of the dataset, where 0\% uses CIFAR$_{10}$ dataset to establish representer points.
For 10\%, 1\%, and 0.1\% sampling rates, we compute mean and standard deviation with 10 random trials.
As shown in Table~\ref{tab:sampling}, the number of samples to compute the foreground predictor has no significant effect on performance drops.
In fact, in certain instances, it even produces better results compared to using the entire dataset.
We also report results of ViT-s based models, such as TokenCut~\cite{tokencut}, ViTOL~\cite{vitol}, and a few-shot method, SPNet~\cite{spnet}.
SPNet uses 5-shots for 200 fine-grained classes on CUB-200-2011 and 10-shots for 1K classes on ImageNet-1K, with object localization annotations, while ViTOL and TokenCut uses a full dataset to train the model.
Even with an extremely small amount of samples in the dataset, our method  surpass SPNet, ViTOL, and TokenCut on ImageNet.

\begin{figure}[t]
    \centering
    \includegraphics[width=0.85\linewidth]{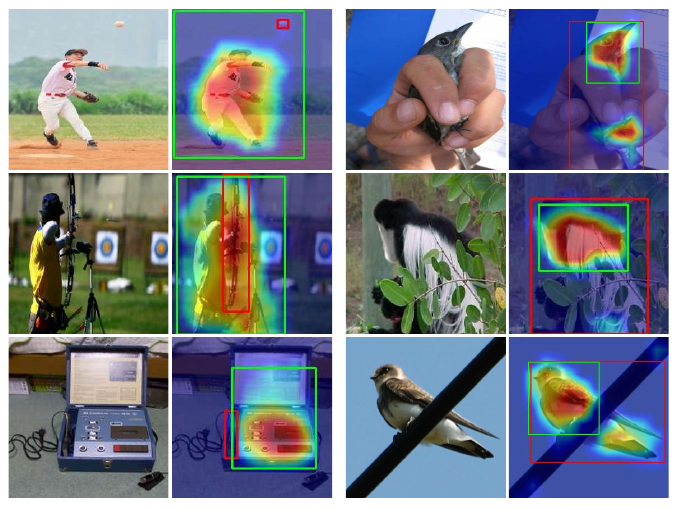}
    \caption{Failure cases of our method on ImageNet-1K (left) and CUB-200-2011 (right) datasets.}
    \label{fig:failurecase}
\end{figure}

\vspace{-0.2cm}
\paragraph{Impact of threshold $\tau$.}
{
We conduct an empirical analysis to show the impact of a threshold $\tau={\|\mathbf{v}\|}/{\|\mathbf{u}\|}$.
As demonstrated by the results in Figure~\ref{fig:taugraph}, we found that the proposed threshold $\tau$ consistently achieved nearly optimal scores, even when tested with various thresholds.
Our defined threshold $\tau$ also consistently performs well across diverse models.
Further detailed theoretical analysis for $\tau$ is provided in \textit{our supplementary material}.}

\vspace{0.2cm}
\subsection{Qualitative Results and Failure Cases}
We illustrate the qualitative results of our method and C$^2$AM in Figure~\ref{fig:result}. 
Both methods use the MoCo v2 model with the ResNet50 backbone.
Without any training on the dataset unlike C$^2$AM, our method successfully localizes objects on various datasets.
Compared with the other method, our activation map shows the more precise area of the object, and activations in the background are suppressed.


We also demonstrate some failure cases in Figure~\ref{fig:failurecase}.
Since our method only relies on features from the pre-trained networks without additional training, it has limited ability to find the small target object particularly in scenarios where multiple objects coexist within the image.
We observe that our method tends to find objects with a dominant size in the image.
Moreover localizing the whole object is challenging in cases where the object is either occluded or fragmented.
These failure cases, however, are not only our limitations but also the limitations of current unsupervised and weakly supervised object localization methods.

\section{Conclusion}
In this paper, we propose a simple and effective unsupervised object localization method without using additional training and supervision.
We leverage the representer point selection in the object localization task, which allows us to explain the decision in terms of the activations of data points.
The proposed method outperforms not only the state-of-the-art self-supervised and unsupervised object localization methods by a significant margin but also surpasses recent weakly supervised and few-shot methods.

\vspace{-0.2cm}
\small{
\paragraph{Acknowledgement.}
This work was supported by the IITP grants (No.2019-0-01842, No.2021-0-02068, No.2022-0-00926) funded by MSIT, the ISTD program (No.20018334) funded by MOTIE, and the GIST-MIT Research Collaboration grant funded by GIST, Korea.
}

{\small
\bibliographystyle{ieee_fullname}
\bibliography{egbib}
}

\appendix
\cleardoublepage

{
   \newpage
       \twocolumn[
        \centering
        \Large
        \textbf{Unsupervised Object Localization with Representer Point Selection}\\
        \vspace{0.5em}\textbf{\textit{-Supplementary Material-}} \\
        \vspace{1.0em}
       ] 
   }


\begin{table*}
\footnotesize
  \centering
    \caption{Comparison between our method and several object localization methods that use the additional classifier, EfficientNetB7~\cite{efficientnet}, in terms of \textit{Top-1}, \textit{Top-5} and \textit{GT-known Loc} on CUB-200-2011 test set and ImageNet-1K validation set. Loc. and Cls. denote the localization and classification backbones, respectively. $\dagger$ indicates MoCo~v2 pre-trained backbone.}
  \begin{tabular}{lccccccccccc}
    \toprule
    \multirow{2}{*}{Method}& \multirow{2}{*}{Loc.} & \multirow{2}{*}{Cls.} & \multirow{2}{*}{$\mathcal{S}$} & \multirow{2}{*}{$\mathcal{T}$} & \multicolumn{3}{c}{CUB-200-2011} & & \multicolumn{3}{c}{ImageNet-1K} \\ \cmidrule{6-8} \cmidrule{10-12}
    & & & & & \textit{Top-1 Loc} & \textit{Top-5 Loc} & \textit{GT-Known} && \textit{Top-1 Loc} & \textit{Top-5 Loc} & \textit{GT-Known} \\
    \midrule
    \multicolumn{3}{l}{\textit{Weakly supervised method}} & & & & & & & & &\\
    {SPOL $_{\text{'}21}$~\cite{SPOL}} & ResNet50 & EfficientNetB7 & {\cmark}  & {\cmark} & {80.12} & {93.44} & {96.46} & & {59.14} & {67.15} & {69.02} \\
    \hdashline[1pt/1pt]
    \multicolumn{3}{l}{\textit{Self-supervised methods}} & & & & & & & & &\\ 
    {PSOL $_{\text{'}20}$~\cite{PSOL}} & DenseNet161 & EfficientNetB7 & {\xmark}  & {\cmark} & {80.89} & {89.97} & {91.78} & & {58.00} & {65.02} & {66.28} \\
    {C$^2$AM $_{\text{'}22}$~\cite{CCAM}} & DenseNet161 & EfficientNetB7 & {\xmark}  & {\cmark} & {81.76} & {91.11} & {92.88} & & {59.56} & {67.05} & {68.53} \\
    \hdashline[1pt/1pt]
    \textit{w/o finetuning} \\
    {Ours} & ResNet50$^{\dagger}$ & EfficientNetB7 & {\xmark}  & {\xmark} & {{\textbf{84.90}}} & {{\textbf{94.74}}} & {{\textbf{96.67}}} & & {\textbf{60.17}} & {\textbf{67.87}} & {\textbf{69.30}} \\
    \bottomrule
  \end{tabular}
  \label{tab:efficient}
\end{table*}

\begin{table}
\footnotesize
  \centering
    \caption{\footnotesize{Comparison between the proposed method and the state-of-the-art weakly supervised object localization methods in terms of \textit{MaxBoxAccV2} on CUB-200-2011 and ImageNet-1K.}}
    \scalebox{0.95}{
  \begin{tabular}{lccc}
    \toprule
    Methods & Backbone & CUB-200-2011 & ImageNet-1K \\
    \midrule
    \multicolumn{3}{l}{\textit{WSOL methods}} \\
    BGC $_{\text{'}22}$~\cite{bgc} & ResNet50 & 75.90 & 68.70 \\
    CREAM $_{\text{'}22}$~\cite{cream} & ResNet50 & 73.50 & 67.40 \\
    DAOL $_{\text{'}22}$~\cite{DAWSOL} & ResNet50 & 69.87 & 68.23 \\
    BagCAM $_{\text{'}22}$~\cite{Bag} & ResNet50 & 84.88 & 69.97 \\
    ViTOL $_{\text{'}22}$~\cite{vitol} & ViT-S & 73.17 & \underline{70.47} \\
    SCM $_{\text{'}22}$~\cite{SCM} & ViT-S & \textbf{89.90} & - \\
    \hline
    \multicolumn{3}{l}{\textit{Self-supervised methods}} \\
    C$^2$AM $_{\text{'}22}$~\cite{CCAM} & ResNet50 & 83.80 & 66.80 \\
    \hline
    \multicolumn{3}{l}{\textit{w/o finetuning }} \\
    MoCo~v2~\cite{mocov2} + Ours & ResNet50 & 87.26 & 66.38 \\
    DINO~\cite{dino} + Ours & ViT-S & \underline{88.83} & \textbf{73.04} \\
    \bottomrule
  \end{tabular}
  }
  \label{tab:maxboxaccv2}
  \vspace{-0.2cm}
\end{table}
\section{Further Analysis of Threshold $\tau$}
In this supplementary section, we offer theoretical and empirical support to determine the threshold value, $\tau$.
As defined in Eq.~(5) of our original manuscript, $\mathbf{w}^*$, which is used as a foreground predictor, can be rewritten with the sample global importance $\alpha$ from Eq.~(9) as follows:
\begin{align}
    \mathbf{w}^*&=\sum_{n=1}^{N_D}\sum_{i=1}^{HW}\alpha_{n,i}\mathbf{\hat{f}}_{n,i}=\sum_{i=1}^{{N}}\alpha_{i}\mathbf{\hat{f}}_{i}\\
    &= \frac{1}{C}\sum_{i=1}^{N}(\|\mathbf{f}_{i}\| - \tau) \mathbf{\hat{f}}_{i}\\
    &= \frac{1}{C}\underbrace{\sum_{i=1}^{N} \mathbf{f}_{i}}_{\mathbf{v}} - \frac{\tau}{C}\underbrace{\sum_{i=1}^{N} \mathbf{\hat{f}}_{i}}_{\mathbf{u}},
    \label{w}
\end{align}
where $\mathbf{f}_{i}$ is a feature vector, $\mathbf{\hat{f}}_i=\frac{\mathbf{f}_{i}}{||\mathbf{f}_{i}||}$, $N=N_DHW$ for simplicity and $C$ denotes a constant.
In Eq.~\eqref{w}, $\tau$ is a threshold that is used to determine soft pseudo labels for the training examples using the norm of the feature vector.

A straightforward approach to determine the threshold is to calculate the expected value of feature vector norms across the training set, given by $\tau=E[||\mathbf{f}_i||]~=~{\sum_i^N||\mathbf{f}_i||/N}$.
However, using a uniform probability distribution for expected value calculations does not provide information on the directions and similarities between feature vectors.
Therefore, we propose a joint probability distribution that considers the correlations among feature vectors to compute the expected value.
Let us denote two independent random variables, ${X}$ and ${X^\prime}$, which share the sample space of feature vector norms, and ${XX^\prime}$ is a joint random variable.
To utilize the relationships between all pairs of feature vectors, we employ the cosine similarity to compute the joint probability mass function of ${XX^\prime}$.
The joint probability mass function of ${XX^\prime}$ is then expressed as follows:
\begin{align}
    P({X}=\|\mathbf{f}_i\|, {X^\prime}=\|\mathbf{f}_j\|) \propto  \mathbf{\hat{f}}_i^\top\mathbf{\hat{f}}_j,
\end{align}
and the expectation of ${XX^\prime}$ is given by
\begin{align}
    E[XX^\prime]&=\sum_{i=1}^{{N}}\sum_{j=1}^{{N}} \frac{\|\mathbf{f}_{i}\|\|\mathbf{f}_{j}\|P({X}=\|\mathbf{f}_i\|, {X^\prime}=\|\mathbf{f}_j\|)}{\sum_{i^\prime=1}^{{N}}\sum_{j^\prime=1}^{{N}}P({X}=\|\mathbf{f}_i^\prime\|, {X^\prime}=\|\mathbf{f}_j^\prime\|)}\\ &=\sum_{i=1}^{{N}}\sum_{j=1}^{{N}}\|\mathbf{f}_{i}\|\|\mathbf{f}_{j}\|\frac{\mathbf{\hat{f}}_{i}^{\top}\mathbf{\hat{f}}_{j}}{\sum_{i^\prime=1}^{{N}}\sum_{j^\prime=1}^{{N}}\mathbf{\hat{f}}_{i^\prime}^{\top}\mathbf{\hat{f}}_{j^\prime}},
\end{align}
where $\mathbf{\hat{f}}_i^\top\mathbf{\hat{f}}_j>0, \forall i,j$, because the last layer of pre-trained encoder $\Phi(\cdot)$ contains a ReLU operation. 

Let us denote $\tau=E[{X}]$, and then the expected value of a jointly distributed discrete random variables of two independent random variables is given by the product of the expected values of two random variables as follows:
\begin{align}
   \tau^2&=(E[{X}])^2=E[{X}]E[{X^\prime}]=E[{XX^\prime}]\\ 
   &=\sum_{i=1}^{{N}}\sum_{j=1}^{{N}}\|\mathbf{f}_{i}\|\|\mathbf{f}_{j}\|\frac{\mathbf{\hat{f}}_{i}^{\top}\mathbf{\hat{f}}_{j}}{\sum_{i^\prime=1}^{{N}}\sum_{j^\prime=1}^{{N}}\mathbf{\hat{f}}_{i^\prime}^{\top}\mathbf{\hat{f}}_{j^\prime}}\\
  &=\frac{\sum_{i=1}^{{N}}\sum_{j=1}^{{N}}\mathbf{{f}}_{i}^{\top}\mathbf{{f}}_{j}}{\sum_{i^\prime=1}^{{N}}\sum_{j^\prime=1}^{{N}}\mathbf{\hat{f}}_{i^\prime}^{\top}\mathbf{\hat{f}}_{j^\prime}},
  \label{tausquare}
\end{align}
where the denominator and numerator in Eq.~\eqref{tausquare} can be expressed by $\mathbf{u}$ and $\mathbf{v}$ as in Eq.~\eqref{w} as follows:
\begin{align}
    \sum_{i=1}^{{N}}\sum_{j=1}^{{N}}\mathbf{\hat{f}}_{i}^{\top}\mathbf{\hat{f}}_{j}&=(\sum_{i=1}^{{N}}\mathbf{\hat{f}}_{i})^{\top}\sum_{i=1}^{{N}}\mathbf{\hat{f}}_{i} \\ 
    &=\|\sum_{i=1}^{{N}}\mathbf{\hat{f}}_i\|^2=\|\mathbf{u}\|^2,\\
    \sum_{i=1}^{{N}}\sum_{j=1}^{{N}}\mathbf{f}_{i}^{\top}\mathbf{f}_{j}&=\|\sum_{i=1}^{{N}}\mathbf{f}_i\|^2=\|\mathbf{v}\|^2.
\end{align}
Therefore, $\tau$ is computed as follows:
\begin{align}
    \tau=\frac{\|\mathbf{v}\|}{\|\mathbf{u}\|}.
    \label{eq:tau}
\end{align}


\section{Additional WSOL Results}
\paragraph{Advanced Classifier}
In comparing our method with other weakly supervised object localization methods~\cite{SPOL,PSOL,CCAM} that utilize more advanced classifiers, such as  EfficientNetB7~\cite{efficientnet}, we also evaluate our method using EfficientNetB7 for classification in a weakly supervised setting.
As shown in Table~\ref{tab:efficient}, our method outperforms other self-supervised methods which utilize much deeper networks, such as DenseNet161~\cite{densenet}, instead of ResNet50~\cite{resnet}, by significant margins.
Furthermore, our method exhibits superior performances compared to the weakly supervised method~\cite{SPOL} which relies on explicit class labels for training, while our method and self-supervised methods solely use pre-trained classification networks for classification.
\paragraph{\textit{MaxBoxAccv2}}
In catering to various demands for localization accuracy, \cite{evalright} proposed evaluating WSOL methods through \textit{MaxBoxAccV2}.
\textit{MaxBoxAccV2} is calculated by averaging the \textit{MaxBoxAcc} performance across various IoU threshold $\delta \in \{0.3, 0.5, 0.7\}$.
As shown in Table~\ref{tab:maxboxaccv2}, our method surpasses other self-supervised and weakly supervised methods on ImageNet.
In the CUB-200-2011 dataset as well, our approach achieves performance with negligible differences from the state-of-the-art, independent of the architecture.

\begin{table}
\footnotesize
  \centering
    \caption{\footnotesize{Comparison of the performance of our method between Moco~v2 and supervised pre-trained ResNet50 on UOL setup.}}
    \scalebox{0.95}{
  \begin{tabular}{lccccc}
    \toprule
    Pre-training & CUB & Cars & Aircraft & Dogs & ImageNet \\
    \midrule
    Supervised & 88.45 & 96.98 & 98.47 & 89.53 & 63.22 \\ 
    MoCo~v2 & \textbf{96.67} & \textbf{99.69} & \textbf{98.71} & \textbf{95.07} & \textbf{66.89} \\
    \bottomrule
  \end{tabular}
  }
  \label{tab:sslandsl}
\vspace{-0.2cm}
\end{table}

\section{Advantage of SSL Pre-trained Backbone}
We present the results of both supervised and self-supervised pre-trained models in Table~\ref{tab:sslandsl}.
Interestingly, we found that the results of the supervised model were inferior to those of the self-supervised model.
This disparity can be linked to a well-established challenge in object localization with class-level supervision~\cite{ADL,HaS,AE}.
Class-level supervised models often concentrate mainly on the most discriminative parts, as they are trained to learn features that have a substantial impact on classification.
In the context of our method, which does not involve fine-tuning the model, this issue becomes more pronounced.
To further illustrate this phenomenon, we include examples in Fig.~\ref{fig:activation}.
Here, the supervised model activates only the most visually prominent features, while the self-supervised model demonstrates a more comprehensive ability to localize the entire object.

\begin{figure}[t]
\begin{center}
   \includegraphics[width=1\linewidth]{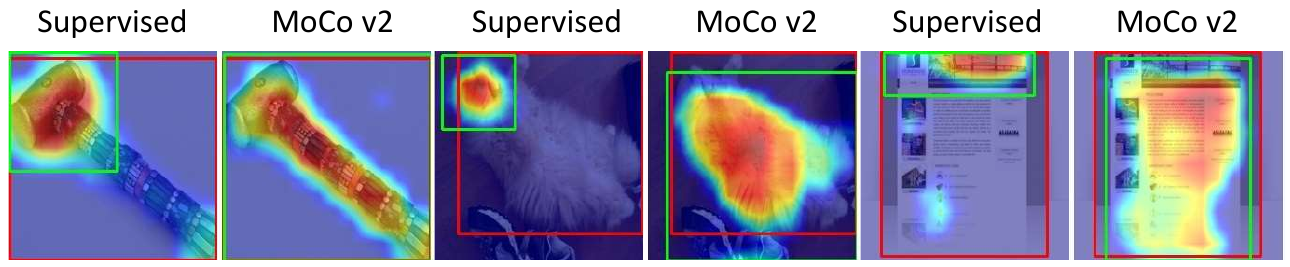}
\end{center}
\vspace{-0.4cm}
   \caption{\footnotesize{Visualization of activation maps using the supervised and self-supervised (MoCo v2) pre-trained models.}}
\vspace{-0.5cm}
\label{fig:activation}
\end{figure}

\begin{figure*}
    \centering
    \includegraphics[width=1\linewidth]{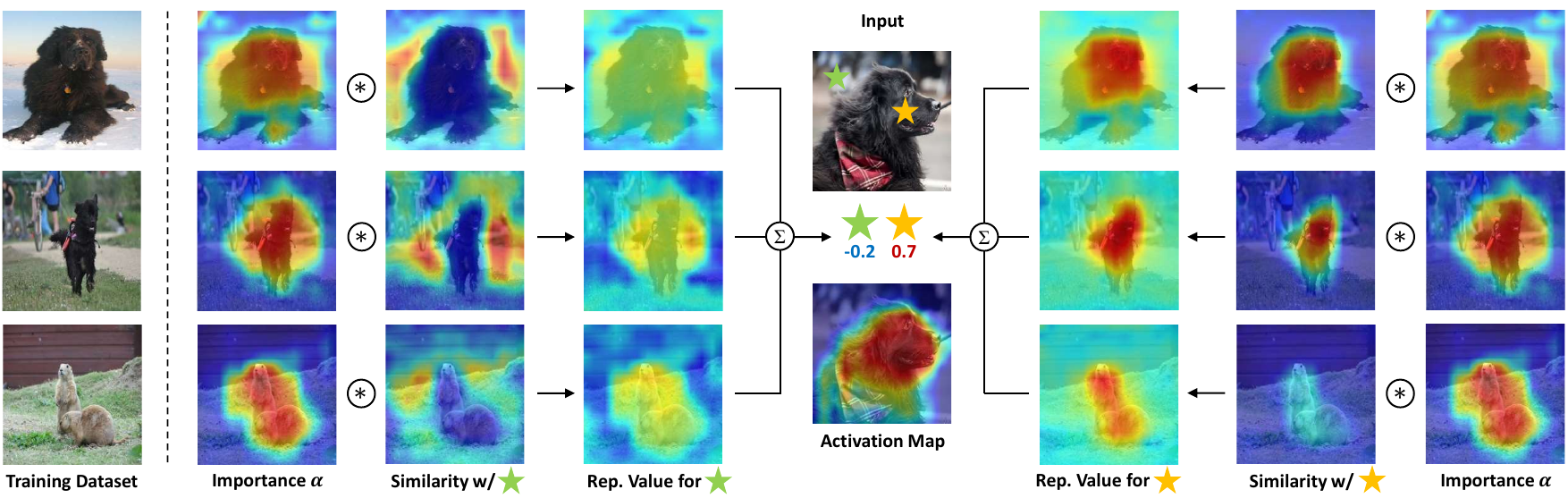}
    \includegraphics[width=1\linewidth]{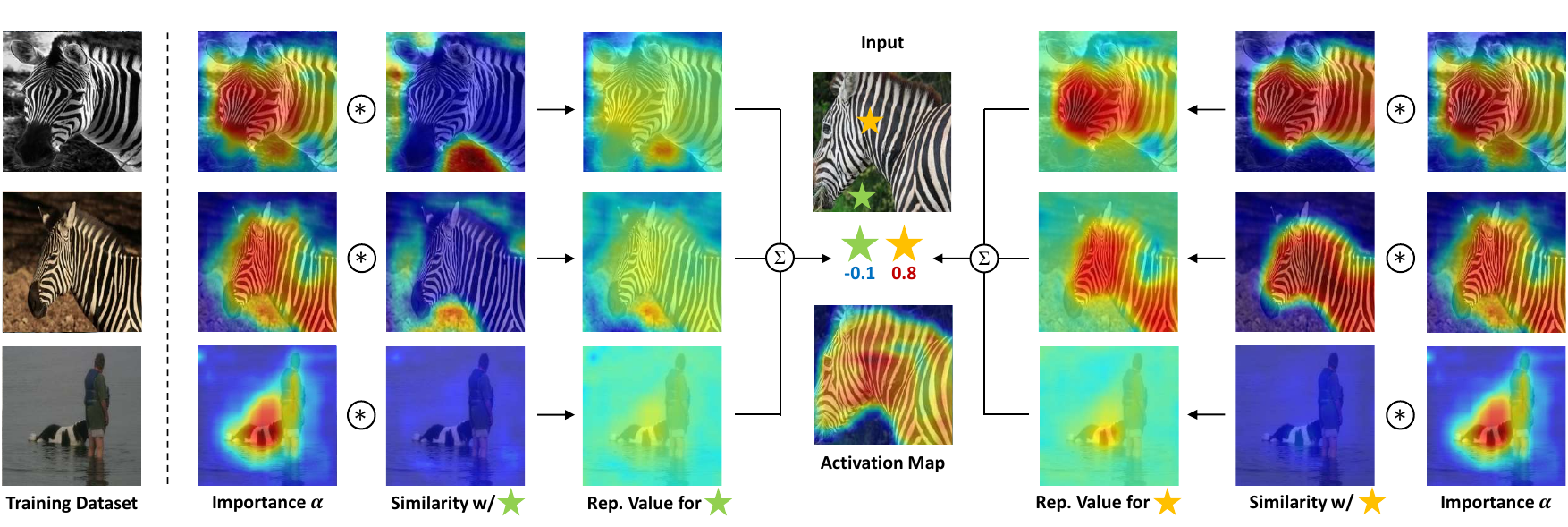}
    \caption{Illustration of how our method computes activation maps using two example points, yellow and green stars. Both importance maps and representer value maps are normalized to be centered at zero, unlike similarity maps' min-max normalization. Hence, red and blue regions denote positive and negative representer points, respectively. Green or yellow colored regions indicate very small absolute values of the representer value.}
    \label{fig:cases}
\end{figure*}

\section{More Qualitative Results}
We also include further qualitative results to illustrate the operating process of our method for selecting representer points, as depicted in Figure~3 of the main manuscript.
As shown in Figure~\ref{fig:cases}, we present examples that highlight global sample importance, the similarity between features, and representer values for given points within an image.
These examples reveal that representer values tend to escalate when both the feature similarity and the importance $\alpha$ of each training example are pronounced.
In the visualizations of representer value maps, red regions indicate excitatory points for the foreground prediction, while blue regions indicate inhibitory points.
By fostering a comprehensive understanding of the model's predictions, it provides valuable insights into the reasoning behind the model's specific predictions and exclusions.

\section{Limitations and Social Impacts}
Our method does not rely on ground-truth annotations, which reduces the risk of bias, but it increases the likelihood of errors in object localization when compared to supervised methods.
In addition, our method shares common limitations with other unsupervised, self-supervised, or weakly supervised methods, such as difficulties in detecting and recognizing rare, small, or complexly appearing objects including objects with similar textures or shapes and those set against cluttered backgrounds.
Additionally, since our approach utilizes training examples, it carries a potential risk of privacy violations if the dataset is not meticulously curated.
However, despite these limitations, we believe our method offers a unique advantage: it provides explainability about how it discovers objects.
This ability sets our approach apart from other methods and adds to its appeal.

\end{document}